\documentclass[letterpaper, 10 pt, conference]{ieeeconf}  

\IEEEoverridecommandlockouts                              

\usepackage{amsmath,bm,amssymb,amsfonts}

\usepackage{svg}
\usepackage{algorithmic}
\usepackage{graphicx}
\usepackage{textcomp}
\usepackage{xcolor}
\usepackage{subcaption}
\usepackage{tabularx}
\usepackage[acronym,nomain,nonumberlist]{glossaries}
\usepackage{cite}

\usepackage{lipsum}

\newcommand\blfootnote[1]{%
  \begingroup
  \renewcommand\thefootnote{}\footnote{#1}%
  \addtocounter{footnote}{-1}%
  \endgroup
}

\makeatletter
\let\NAT@parse\undefined
\makeatother
\usepackage{hyperref}  

\begin{document}

\newacronym{vae}{VAE}{Variational Autoencoder}
\newacronym{ldm}{LDM}{Latent Diffusion Model}
\newacronym{ddm}{DDM}{Denoising Diffusion Model}
\newacronym{cvae}{CVAE}{Conditional Variational Autoencoder}
\newacronym{gldm}{GraspLDM}{Grasp Latent Diffusion Models}
\newacronym{emd}{EMD}{Earth Mover's Distance}
\newacronym{mrp}{MRP}{Modified Rodrigues Parameters}
\newacronym{edm}{EDM}{Elucidated Diffusion Model}
\newacronym{sde}{SDE}{Stochastic Differential Equations}
\newacronym{ode}{ODE}{Ordinary Differential Equations}
\newacronym{sgm}{SGM}{Score-based Generative Model}
\newacronym{ddpm}{DDPM}{Denoising Diffusion Probabilistic Models}
\newacronym{elbo}{ELBO}{Evidence Lower Bound}
\newacronym{ddim}{DDIM}{Denoising Diffusion Implicit Model}
\newacronym{6dof}{6-DoF}{six degrees-of-freedom}
\newacronym{slam}{SLAM}{Simultaneous Localization and Mapping}
\newacronym{cnn}{CNN}{Convolutional Neural Networks}
\newacronym{nerf}{NeRF}{Neural Radiance Fields}
\newacronym{3dgs}{3DGS}{3D Gaussian Splatting}
\title{\LARGE \bf
Object-centric Reconstruction and Tracking of Dynamic Unknown Objects Using~\acrlong{3dgs}}

\vspace{-1.0\baselineskip}
\author{Kuldeep R Barad$^{1,2}$, Antoine Richard$^{1}$, Jan Dentler$^{2}$, \\ Miguel Olivares-Mendez$^{1}$ and Carol Martinez$^{1}$
\thanks{$^{1}$Space Robotics Research Group (SpaceR), Interdisciplinary Centre for Security, Reliability and Trust (SnT), University of Luxembourg, Luxembourg.
        {\tt\small kuldeep.barad@uni.lu}}%
\thanks{$^{2}$Redwire Space Europe, Luxembourg
        }%
}

\makeatletter
\let\@oldmaketitle\@maketitle
\renewcommand{\@maketitle}{\@oldmaketitle
    \centering
  \includegraphics*[width=0.95\linewidth]
    {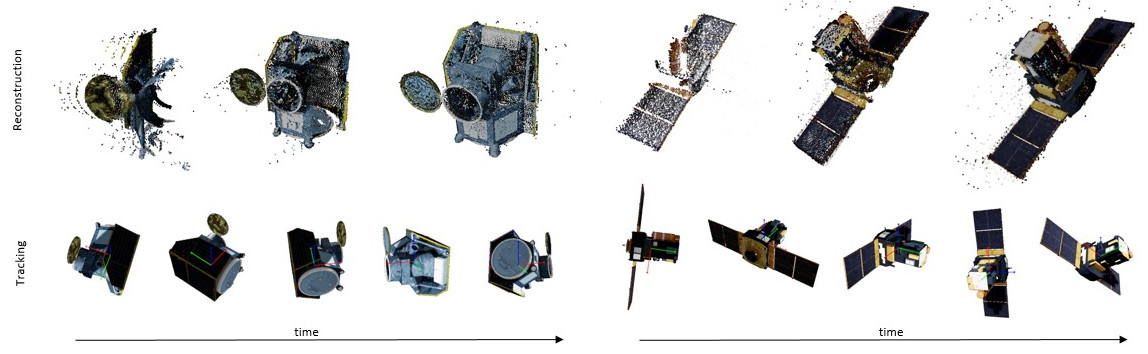}
    \label{fig:header}
    \captionof{figure}{Incremental reconstruction and tracking of CHEOPS (left) and SOHO (right) spacecraft from a sequence of simulated images without prior training using object-level 3D Gaussian representations and splatting-based differentiable rendering.}}
    \vspace{-1.0\baselineskip}
\makeatother

\maketitle

\thispagestyle{empty}
\pagestyle{empty}

\begin{abstract}
Generalizable perception is one of the pillars of high-level autonomy in space robotics. 
Estimating the structure and motion of unknown objects in dynamic environments is fundamental for such autonomous systems. 
Traditionally, the solutions have relied on prior knowledge of target objects, multiple disparate representations, or low-fidelity outputs unsuitable for robotic operations. 
This work proposes a novel approach to incrementally reconstruct and track a dynamic unknown object using a unified representation- a set of 3D Gaussian blobs that describe its geometry and appearance. 
The differentiable~\acrshort{3dgs} framework is adapted to a dynamic object-centric setting. 
The input to the pipeline is a sequential set of RGB-D images. 3D reconstruction and~\acrshort{6dof} pose tracking tasks are tackled using first-order gradient-based optimization.
The formulation is simple, requires no pre-training, assumes no prior knowledge of the object or its motion, and is suitable for online applications.
The proposed approach is validated on a dataset of 10 unknown spacecraft of diverse geometry and texture under arbitrary relative motion. 
The experiments demonstrate successful 3D reconstruction and accurate~\acrshort{6dof} tracking of the target object in proximity operations over a short to medium duration. 
The causes of tracking drift are discussed and potential solutions are outlined. 
\end{abstract}

\section{Introduction}
\blfootnote{Supplementary materials are available at \url{https://kuldeepbrd1.github.io/projects/oc-3dgs/}} 
Generalizable perception systems are crucial for terrestrial and space-borne robotic systems that operate autonomously in dynamic and unstructured environments.
In such use cases, the robot needs to reason about the geometric and physical properties of the objects around it for safe navigation and interaction. In challenging applications like on-orbit servicing, robotic interaction, and debris removal, prior knowledge of the target object may not be available or reliable.  
These object properties must be inferred online using noisy and partial observations from sensors. Due to lower size, weight, power, and cost (SWaP-C), vision sensors are a cost-effective choice for robotic systems. However, extracting object-level properties like geometric structure and motion online from image pixels is challenging. Therefore, we are interested in structure and motion recovery methods that offer: 1) efficient object-level representations and 2) generalization to an open set of objects. 


Extraction of object-level structure and relative motion from visual observations concerns two fundamental problems in computer vision- 3D reconstruction and~\acrfull*{6dof} pose tracking.
The two problems are generally dealt with separately, while the solutions often use different internal representations and assumptions.
For instance, the state-of-the-art methods for scene reconstruction in terrestrial applications use Neural Radiance Field (NeRF) representations~\cite{mildenhall2021nerf}. NeRF-based methods usually assume a static scene and the availability of relative camera poses.
On the other hand,~\acrshort{6dof} pose estimation and tracking for unknown objects is addressed with large pre-trained neural networks.
Many state-of-the-art approaches require a 3D model of the object as a template~\cite{labbe2022megapose}.
Furthermore, a downstream robotic task like grasping may use another neural network that takes RGB-D images as inputs to infer grasps.
All these elements require separate datasets, isolated pre-training(s), and parameter tuning.

We focus on the problem of vision-based uncooperative proximity operations with an unknown target spacecraft. Recent works have addressed robustness challenges in pre-training~\acrfull*{cnn} models for a single known object~\cite{pauly2023survey}. 
Tracking is constrained to natural orbital motion~\cite{cassinis2021evaluation} and 3D reconstruction is limited to a set of primitive shapes~\cite{park2024rapid}. In this work, we tackle the 3D reconstruction and~\acrshort{6dof} pose tracking of an unknown object incrementally with a unified underlying representation. Our approach assumes no prior knowledge about the object's structure or motion. 
We represent an object with a set of points and 3D Gaussian blobs centered at these points with color and opacity properties. 
Our goal is to fit this object-level representation to an incremental sequence of RGB-D observations. 
To accomplish this, we use a differentiable rendering framework based on~\acrfull*{3dgs}~\cite{kerbl20233d} to model the image formation process.
The forward map of the differentiable renderer projects the object representation and a camera pose to a rasterized image. 
The backward map of the renderer propagates the gradients from a rendered image to the object and camera parameters. 
3D reconstruction and~\acrshort{6dof} pose tracking can be accomplished by refining the object and camera pose parameters using gradient-based optimization.
The gradient computation can be implemented efficiently using explicitly derived gradients making the optimization process computationally efficient for online use. 
Unlike neural implicit representations~\cite{mildenhall2021nerf}, the forward map of the rendering based on~\acrshort{3dgs} process is significantly faster~\cite{kerbl20233d}. 
Our approach requires no data generation or pre-training. The input is a stream of RGB-D images of an unknown object in unconstrained relative motion. 
Through a simple rendering loss, we optimize the object representation and the camera pose alternately in 120 and 80 steps respectively. 
We test our approach on synthetic image sequences of 10 unique spacecraft models from ESA's science fleet~\cite{esascifleet}. 
We demonstrate the effectiveness of our approach on these long video sequences, highlight limitations, and provide recommendations for future development toward reliable and generalizable perception of dynamic unknown objects. 

In summary, our work makes the following contributions: \textbf{(1) Development of an efficient object-centric framework for incremental reconstruction and tracking of unknown objects using~\acrshort{3dgs}.} The original work~\cite{kerbl20233d} addresses offline scene reconstruction from posed images. Our work relaxes the need for a static world frame or camera poses and proposes an object-centric framework suitable for online deployment. \textbf{(2) Demonstration of high-fidelity reconstruction and~\acrshort{6dof} pose tracking of dynamic unknown spacecraft}. To the best of our knowledge, we provide the first approach to tackle this problem without assuming prior knowledge about the target and its motion.

\section{Related Work}

\subsection{Pose estimation and tracking of uncooperative spacecraft}

~\acrshort{6dof} pose estimation and tracking for cooperative and uncooperative spacecraft are important problems for proximity operations in orbit. Early works investigated a host of sensor and algorithmic choices while highlighting the need for a cost-effective and robust system~\cite{opromolla2017review}. Recently, the focus has shifted towards vision-based pose estimation methods due to the lower SWaP-C of vision sensors. Importantly, the progress was driven by rapidly evolving models and methods in computer vision. Early works~\cite{naasz2009hst,d2014pose} utilized conventional image features that either extract parts of the structures like edges~\cite{canny1986computational} and corners~\cite{harris1988combined} or invariant visual descriptors like SIFT~\cite{lowe1999object}. Here, the pose estimation performance relied on feature extraction and robust correspondence matching to a known 3D model. Due to the challenging visual environment, the methods lack robustness to the challenging visual environment in orbit. Subsequently, \acrshort{cnn}s demonstrated superior~\acrshort{6dof} pose estimation results on representative scenarios~\cite{sharma2020neural}. However, these networks require extensive pre-training which is done with synthetic data. Using synthetic data for pre-training a neural network presents the problem of performance transfer across the domain gap between the synthetic and real image formation processes. Consequently, high-quality datasets~\cite{park2022speed+}, image augmentations~\cite{black2021real} and domain adaptation strategies~\cite{perez2022spacecraft,park2023online} have been proposed to address this issue. Simultaneously, these learned models for pose estimation were extended to state tracking assuming natural motion~\cite{cassinis2021evaluation, park2023adaptive}. Overall, these approaches require exhaustive and iterative cycles of dataset curation, pre-training, and ground testing for every target. Moreover, unmodeled artifacts like structural damage or material degradation remain challenging for these methods relying on the exact 3D model.

The requirement of the target model knowledge was relaxed in~\cite{park2024rapid} which proposed a fast, one-shot primitive reconstruction of the target object. The shape is approximated by a fixed set of superquadrics per image. Despite being fast, the method only recovers the shape of the target and not the scale. Consequently, only a rotation is estimated per image. It also depends on a prior learned from a small dataset of objects. These attributes imply that it suffers from overfitting, viewpoint ambiguities of primitives, and a lack of fidelity. As opposed to these methods, our approach does not require pre-training and considers no apriori information about the object or its relative motion. The object is reconstructed online using a 3D Gaussian representation that simultaneously enables efficient pose tracking in the loop. Our approach applies more generally to any unknown object and can be used for other on-board applications such as robotic manipulation.
\begin{figure*}[th]
     
     \begin{subfigure}[b]{0.65\textwidth}
         
         \includegraphics[width=\textwidth]{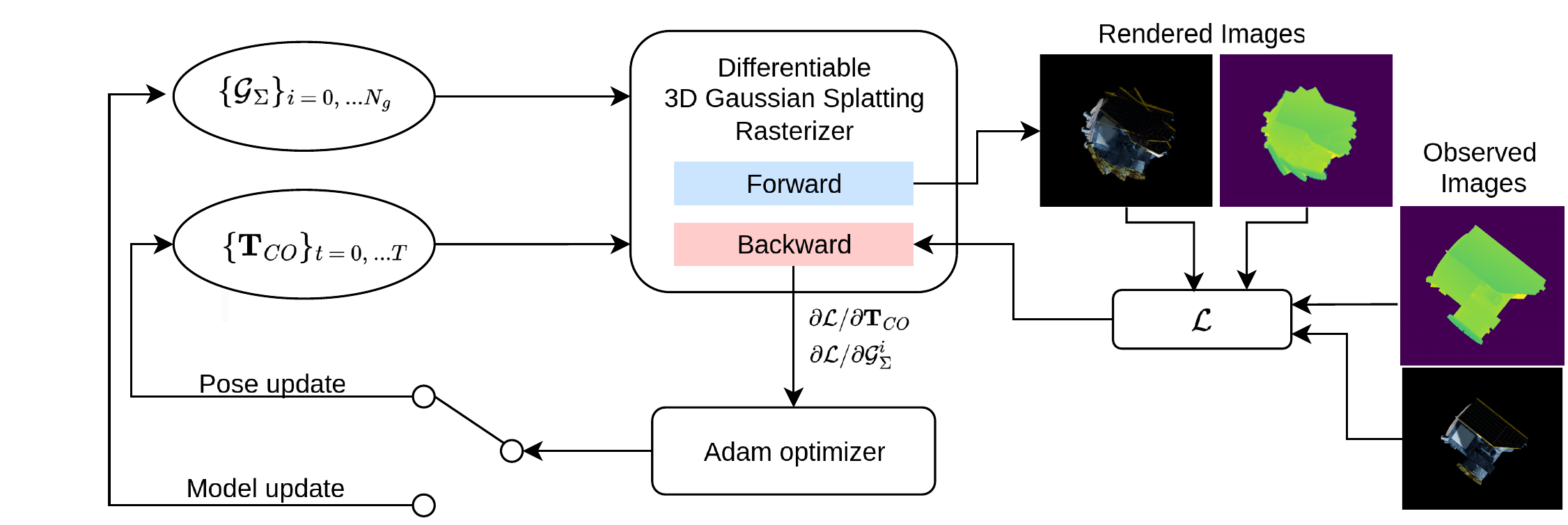}
         \caption{}
         \label{fig:method}
     \end{subfigure}
     \hfill
     \begin{subfigure}[b]{0.32\textwidth}
         \centering
         \includegraphics[width=\textwidth]{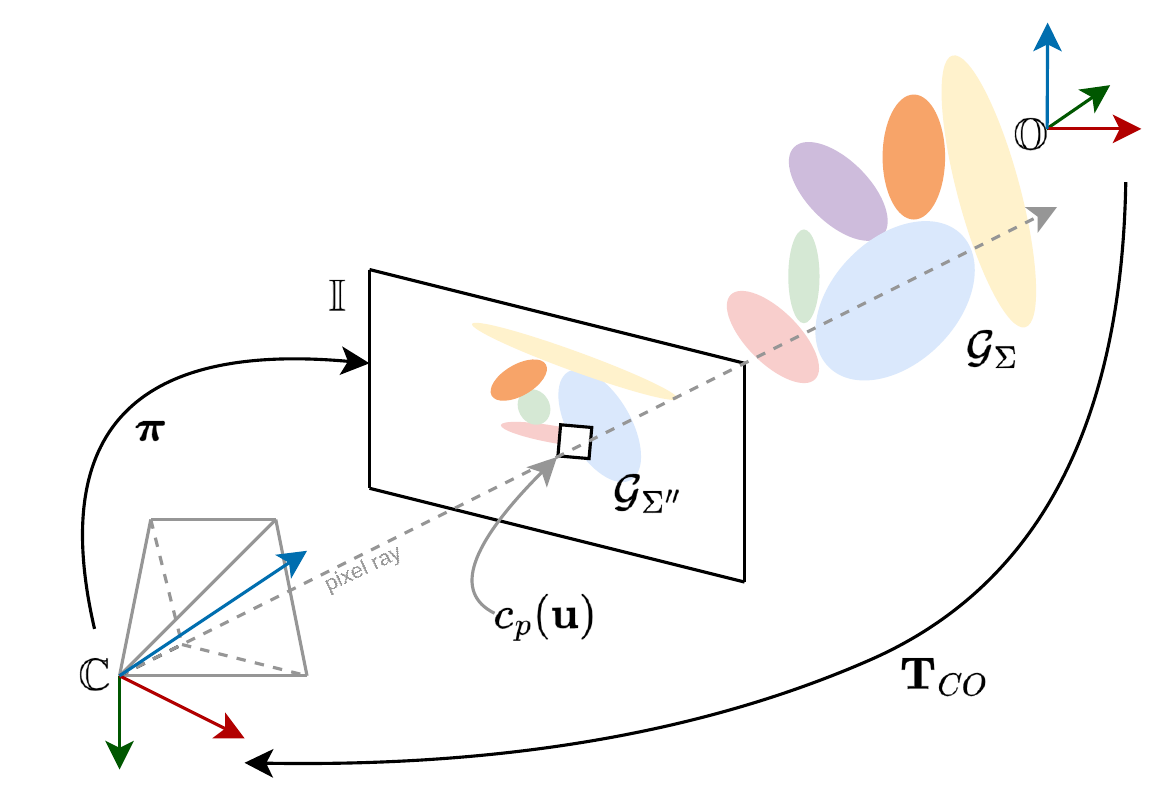}
         \caption{}
         \label{fig:splatting}
     \end{subfigure}
        \caption{(a) \textbf{Methodology}: We use differentiable rendering of 3D Gaussians as the core of our incremental reconstruction and tracking pipeline. We refine the camera pose or the object Gaussians using first-order gradient-based optimization by propagating the gradients backward through the rendering process. (b) \textbf{3D Gaussian Splatting}: Illustration of projecting 3D Gaussians $\mathcal{G}_{\Sigma}$ to 2D Gaussian splats $\mathcal{G}_{\Sigma''}$.}
        \label{fig:three graphs}
        \vspace{-1.0\baselineskip}
\end{figure*}
\subsection{SLAM}
The problem of incremental reconstruction and tracking, addressed in this work, closely relates to the broader research on \acrshort*{slam}. However, SLAM approaches most commonly assume a static world and aim to map and localize a moving camera within the scene~\cite{engel2017direct, forster2016svo}. Sparse map recovery can achieve relatively fast localization of a moving camera. However, the sparse map is less useful for tasks beyond localization. To address this, dense SLAM approaches~\cite{teed2021droid,izadi2011kinectfusion} produce dense maps with point or voxel-based representation. However, mapping and localization are more difficult, if the objects in the environments move. Dynamic SLAM and object-centric SLAM methods relax the static world assumption by segmenting the object and registering a frame-to-model colored point cloud~\cite{ma2015simultaneous} using iterative closest points~\cite{besl1992method} estimation. These methods often use discrete point cloud representations that can be harder to optimize and may not capture high-fidelity details. Recent advancements in 3D reconstruction have enabled the recovery of high-fidelity models from images using differentiable rendering. In particular, neural fields~\cite{mildenhall2021nerf} can be used to create a dense map of the scene~\cite{sucar2021imap,johari2023eslam}. Neural fields have been used for object-centric reconstruction and tracking with RGB-D vision~\cite{wen2023bundlesdf}. However, neural representations are bottlenecked by expensive sampling along pixel rays as they encode a volume implicitly. On the other hand,~\acrshort{3dgs}~\cite{kerbl20233d} provides a new representation that enables high-speed forward rendering and the benefits of dynamic manipulation of point-like 3D Gaussian primitives. This representation has enabled state-of-the-art performance in dense SLAM~\cite{huang2023photo,matsuki2023gaussian,keetha2023splatam}, for large but static scenes. In this work, we investigate the applicability of Gaussian to object-centric reconstruction and tracking applied to characterizing an unknown resident space object.

\section{Method}

Our approach uses a differentiable rasterizer based on~\acrshort{3dgs}~\cite{kerbl20233d} and a simple photometric loss $\mathcal{L}$ to incrementally optimize the object and pose parameters as shown in Fig.~\ref{fig:method}. 
Intuitively, this method stores an evolving internal model of the object and camera motion to render an expected image. 
Then, the photometric distance or dissimilarity between this image and the observed image represents the gap between the true and the internal model. 
The object and camera parameters are then refined using gradient-based optimization. The input to our pipeline is a sequential set of RGB-D images. We assume the target spacecraft is sufficiently resolved in the image and its RGB appearance is view-independent. 
In the following, we describe each component of our method in detail.

\subsection{Object Representation}
We represent the appearance and geometry of the target object as a set of $N\in \mathbb{N}^+$ 3D Gaussian ellipsoids. Each Gaussian $\mathcal{G}^i$ is parameterized in terms of a 3D center ($\boldsymbol{\mu}^i \in \mathbb{R}^3$) and 3D covariance ($\boldsymbol{\Sigma}^i\in \mathbb{S}^3_+$), evaluated at any point $\mathbf{x}$ as:
\begin{equation}
    {\mathcal{G}}_{\boldsymbol{{\Sigma}}}(\mathbf{x}) = \frac{1}{\sqrt{|\boldsymbol{\Sigma}|}} \exp{\left( -\, \frac{1}{2} \,(\mathbf{x}-\boldsymbol{\mu})^T \, \boldsymbol{\Sigma}^{-1} \,(\mathbf{x}-\boldsymbol{\mu})\right)}
\end{equation}
In addition, the appearance of each Gaussian is parameterized by its color ($\mathbf{c}^i$) and opacity ($o^i$), where $i={0,\,1,\,2\,...\,N}$. 
We assume that the appearance of each Gaussian is independent of the viewing direction and exclude the spherical harmonics coefficients from \cite{kerbl20233d} to speed up rendering. 
Note that this representation is explicit and similar to point-based representations. 
Consequently, Gaussian ellipsoids can be added, removed, and modified dynamically during runtime. Unlike implicit representations, this enables better control over reconstruction fidelity and rendering efficiency.

\subsection{Differentiable rendering}
The forward projection of a set of 3D Gaussians result in 2D splats on the image as illustrated in Fig.~\ref{eq:splatting_rasterization}. 
A 3D Gaussian in the object space ($\mathbb{O}$) can be projected to the camera space ($\mathbb{C}$) by an affine view transformation $\mathbf{T}_{CO}\in SE(3)$, with rotation component $\mathbf{R}_{CO}\in SO(3)$ and translation component $\mathbf{t}_{CO}\in \mathbb{R}^3$. 
The Gaussian representation is convenient as the affine transformation results in another 3D Gaussian given by Eq.~\ref{eq:gaussian_projection_camera}.
\begin{equation}
    \mathcal{G}_{\boldsymbol{\Sigma}} (\mathbf{T}_{CO}^{-1}\mathbf{x}_c)= \frac{1}{|\mathbf{R}_{CO}^{-1}|} \mathcal{G}_{\boldsymbol{\Sigma'}} (\mathbf{x}_c)  \,  \; ;  \;  \boldsymbol{\Sigma'}\,=\,\mathbf{R}_{CO} \, \Sigma \, \mathbf{R}_{CO}^T 
    \label{eq:gaussian_projection_camera}
\end{equation}

For a perspective camera, a camera space Gaussian (Eq. \ref{eq:gaussian_projection_camera}) is transformed to the image space ($\mathbb{I}$) using the perspective projection ($\boldsymbol{\Pi}$), which is not affine. 
However, \cite{zwicker2002ewa} introduces a locally affine assumption around a camera coordinate $\mathbf{x}_c^*$ using first-order Taylor expansion in Eq.~\ref{eq:taylor_expansion}. 
Using this approximation the image space projection is a 2D Gaussian given by Eq.~\ref{eq:image-space-gaussian}.
\begin{align}
    \boldsymbol{\Pi}^* (\mathbf{x}_c) &= \boldsymbol{\Pi}(\mathbf{x}^*_c) \, + \, \mathbf{J}_{\Pi}^*\, (\mathbf{x}_c - \mathbf{x}_c^*) \label{eq:taylor_expansion}\\
    \mathcal{G}_{\boldsymbol{\Sigma'}} \,  ( \boldsymbol{\Pi}^{-1} (\mathbf{u})) \; &=\;\frac{1}{|{\mathbf{J}^*_{\Pi}}^{-1}|} \, \mathcal{G}_{\boldsymbol{\Sigma''}} (\mathbf{u})  \label{eq:image-space-gaussian}\\ 
    \boldsymbol{\Sigma''}\,= \,\mathbf{J}^*_{\Pi} \, \boldsymbol{\Sigma'} \, {\mathbf{J}^*_{\Pi}}^{T} 
 &= \,\mathbf{J}^*_{\Pi}\,\mathbf{R}_{CO} \, \boldsymbol{\Sigma} \, \mathbf{R}_{CO}^T \,{\mathbf{J}^*_{\Pi}}^T \nonumber 
\end{align}
where, ${\mathbf{J}^*_{\Pi}}$ is the jacobian of the projective transform linearized at camera space coordinates $\mathbf{x}_c^*$ and $\mathbf{u}$ are the image space coordinates.

The contribution of each Gaussian to a pixel is dependent on the order in which they appear along a ray cast from that pixel. 
For each pixel, the $K\in \mathbb{N}^+$ overlapping Gaussians are sorted from front to back. 
Then, the intensity or color ($\mathbf{c}_p$) of each 2D pixel location ($\mathbf{u}\in\mathbb{R}^2$) in the image is obtained by alpha-blending the contributions of the $K$ Gaussians as in Eq.~\ref{eq:splatting_rasterization}, following the approximations to the volumetric rendering equations in~\cite{zwicker2002ewa}.
\begin{equation}
    \mathbf{c}_p(\mathbf{u}) \; = \; \sum_{k=0}^{k\leq K} \mathbf{c}^k \alpha^{k} \prod_{j=0}^{k-1} ( 1\,-\,\alpha^j)
    \label{eq:splatting_rasterization}
\end{equation}

\begin{equation*}
    \alpha^k = o^k \mathcal{G}^k_{\boldsymbol{\Sigma''}}(\mathbf{u})
\end{equation*}

where, $\alpha^k$ is the contribution of the $k^{th}$ Gaussian to the pixel intensity. 
This contribution is computed by decaying the assigned opacity ($o^k$) by the 2D Gaussian~($\mathcal{G}_{\boldsymbol{\Sigma''}}$). 

As the 3D location of each Gaussian is known, we can similarly render a depth image using per-point $z$ distance in the camera frame: 
\begin{equation}
    d_p(\mathbf{u}) \; = \; \sum_{k=0}^{k\leq K} z^k \alpha^{k} \prod_{j=0}^{k-1} ( 1\,-\,\alpha^j)
    \label{eq:splatting_rasterization_depth}
\end{equation}

This splatting and blending process is fully differentiable and does not use neural elements. 
The image formation model in Eq.~\ref{eq:splatting_rasterization} is similar to the image formation model in ~\acrlong*{nerf}~\cite{mildenhall2021nerf}. 
However,~\acrshort*{nerf}s implicitly represent free space inside a volume, requiring an expensive sampling process along the ray. 
On the other hand, the forward rendering process for Gaussians automatically ignores free-space regions in 3D along the viewing direction. 
As a result, rendering 3D Gaussians is considerably faster than a ~\acrshort*{nerf}.

\subsection{Reconstruction and Tracking}
Assuming no constraints on the relative motion between the camera and the target object, our goal is to incrementally incorporate new information revealed in an image sequence to reconstruct and track a specific object in the scene. 
Consider the rendering process from the previous section, encapsulated as:
\begin{equation}
    \mathbf{I}_r = \mathcal{R}(\boldsymbol{\theta}, \mathbf{T}_{CO})
\end{equation}
where, $\mathcal{R}$ is the renderer that projects a set of 3D Gaussians representing the object to a rasterized image~($\mathbf{I}_r$). $\boldsymbol{\theta}$ are the object parameters obtained by concatenating the per-point Gaussian parameters: $\boldsymbol{\theta}^i = [\boldsymbol{\mu}^i, \boldsymbol{\Sigma}^i, \mathbf{c}^i, o^i]$ for $i={0,1,2,..., N}$. Since the renderer is fully differentiable, we can compute and propagate gradients from image pixels to the object and camera parameters. We use Adam~\cite{kingma2014adam} to optimize the Gaussian and pose parameters.

To reconstruct the object by optimizing the object parameters $\boldsymbol{\theta}$, we require multi-view information to avoid overfitting the representation on restricted regions of the object. 
On the other hand, to optimize the pose, we only require the information from the current frame to establish the frame-to-model association. 
Consequently, we separate optimization into two stages that are executed in alternate frames of the incoming stream.
During reconstruction, we hold camera parameters constant while optimizing the object parameters. 
The optimization fits the object parameters to a weighted combination of photometric loss for the color image and L1 loss for the depth image:
\begin{align}
    L_{recon} = (1\, -\, \lambda)(L_1)_{color} \;+\; & \lambda (L_{SSIM})_{color} \\ &+ \beta (L1)_{depth} \nonumber
\end{align}

where, $L_1$ is the averaged pixel-wise L1 loss between the observed image and the image with a pose $\mathbf{T}_{CO}^n$ for the $n^{th}$ image. 
$L_{SSIM}$ is the structural similarity index loss based on \cite{wang2004image}. 
Unlike scene-reconstruction applications of Gaussian splatting, it is computationally impractical for our application to optimize the object representation using all the images available. 
Therefore, we manage a fixed window of $W$ keyframes to optimize the object representation. 
Like SLAM pipelines, we aim to select keyframes that are most meaningful for multi-view constraints. 
To accomplish this we use up to $W_{max}$ views with farthest angular distance between them and always include the current and the previous view.

For Tracking, the ground truth pose of the first frame is used to establish a canonical frame, if available. Otherwise, it is initialized with a random rotation and centered at the centroid of the partial point cloud de-projected from the first depth image. For the sake of optimization, we reduce the $SE(3)$ pose to a vector $\mathbf{x}_{CO} = [\mathbf{t}_{CO}, \mathbf{q}_{CO}]^T$, where $\mathbf{q}_{CO} = [q_w,\, q_x,\, q_y, \, q_z]$ is the unit quaternion representing relative orientation. The pose is optimized using:

\begin{align}
    L_{track} = (L_1)_{color} \;+\; \beta (L_1)_{depth} \nonumber
\end{align}

The implementation is done using PyTorch using the 3DGS CUDA rasterizer implementation~\cite{kerbl20233d} to obtain the gradients of the loss with respect to the object parameters- $\partial \mathcal{L}/\partial \boldsymbol{\theta}$. 
The gradients of the loss with respect to the pose parameters $\mathbf{x_{CO}}$ are obtained via auto-differentiation. 
We project the Gaussians to the camera frame outside the CUDA rasterizer which provides the gradient flow from the analytical gradients $\partial \mathcal{L}/\partial \boldsymbol{\theta}$. At each new frame, we initialize the pose estimate by linearly propagating the pose as in Eq.~\ref{eq:forward_propagation}, where $t$ is the time index.
\begin{align}
    \mathbf{t}^{t+1} \, = \, \mathbf{t}^t + (\mathbf{t}^t - \mathbf{t}^{t-1}) \\
     \mathbf{q}^{t+1} \, = \, \mathbf{q}^t \otimes (\mathbf{q}^t \otimes \mathbf{\bar{q}}^{t-1})
     \label{eq:forward_propagation}
\end{align}
where, 
For the tracking iteration, this pose is used as the initial guess and refined through optimization. On the other hand, this extrapolated pose stays fixed for the reconstruction iteration. The operations $\otimes$ and $\mathbf{\Bar{q}}$ are quaternion multiplication and inverse respectively. We use 120 optimization steps for reconstruction and 80 steps for tracking.

\subsection{Initialization, Addition, and Removal of 3D Gaussians}
We use the first depth image to initialize the 3D Gaussians~($\mathcal{G}_{\Sigma}$) in the object space. The centers ($\boldsymbol{\mu^i}$) are centered around the point cloud obtained by de-projecting the depth image and the variances are initialized with 0.001. The color is initialized from the corresponding color image pixel value and the opacity is set to 0.5. For the camera-relative pose, if the ground-truth pose is available, it is provided only in the first frame to set the object reference frame. Otherwise, it can be arbitrarily initialized relative to the point cloud obtained from the first depth image. 

As new regions of the object are revealed, we add more points to represent the previously unseen regions. Instead of having a fixed amount of points fit the new observations, this is advantageous for two reasons. First, the optimization is more convenient as a smaller number and size of steps are required to fit the new points. Second, the fidelity of the local geometry can be maintained. We use the difference between the observed depth image and the rendered depth image to add these points at every new frame. A point is added if the difference between the values in the rendered depth image and the observed depth image is larger than 10\% of the difference between maximum and minimum depth. To project the points to the object frame, we need to use the latest available transformation $\mathbf{T}_{CO}$ between the object and the camera frame. As we add more points, the number of optimization parameters grows proportionally. This slows down the optimization steps. More importantly, the points are added with high density at each frame and may be redundant to the object representation. As a result, we prune points for which opacity drops below 0.6 during reconstruction. Together, this process of addition and removal of points balances the need for fidelity and ease of optimization against the computational complexity.

\section{Experiments}
Our method directly optimizes the object representation and camera pose online. Consequently, no pre-training or large-scale data curation is necessary and it can be tested directly on sequential RGB-D data of any unknown object.

\begin{figure*}[th]

     \begin{subfigure}[b]{0.60\textwidth}
         \centering
         \includegraphics[width=\textwidth]{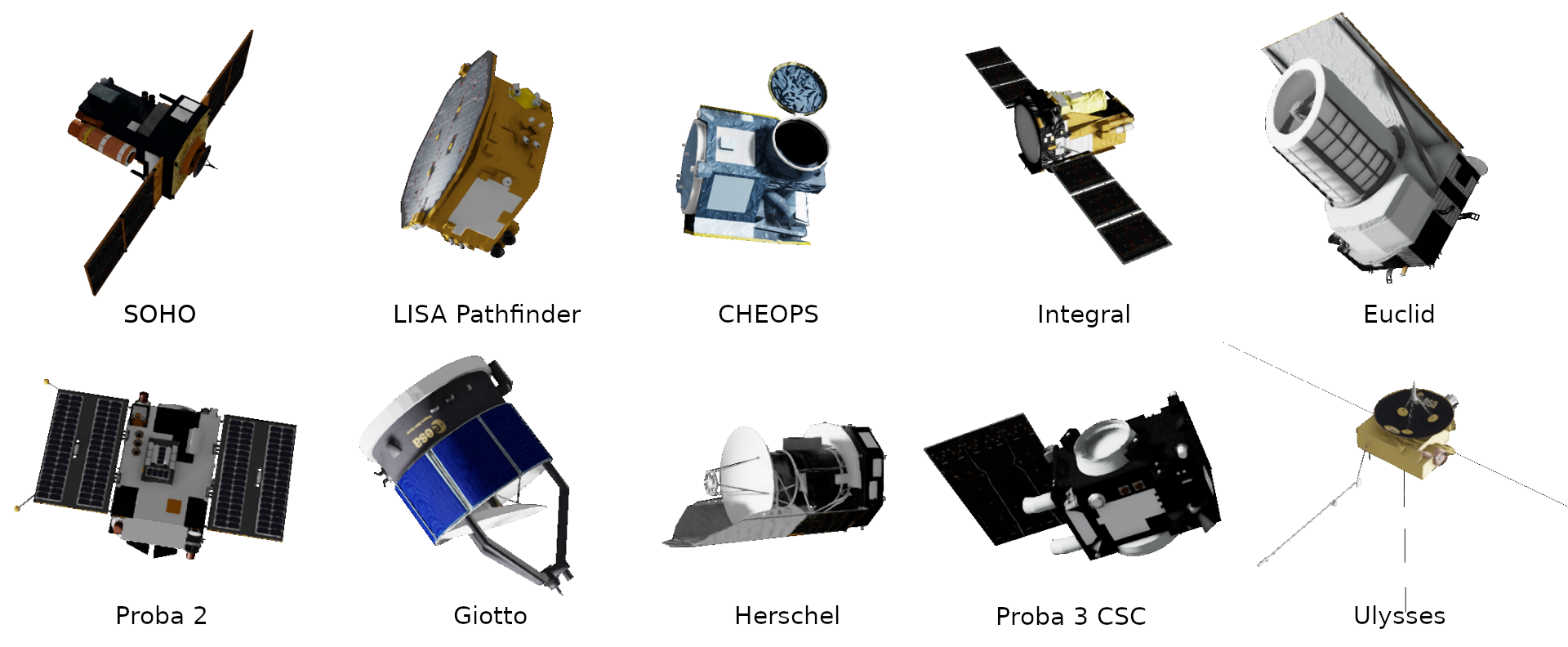}
         \caption{}
         \label{fig:sat_mosaic}
     \end{subfigure}     
     \hfill
     \begin{subfigure}[b]{0.36\textwidth}
         \includegraphics[width=\textwidth]{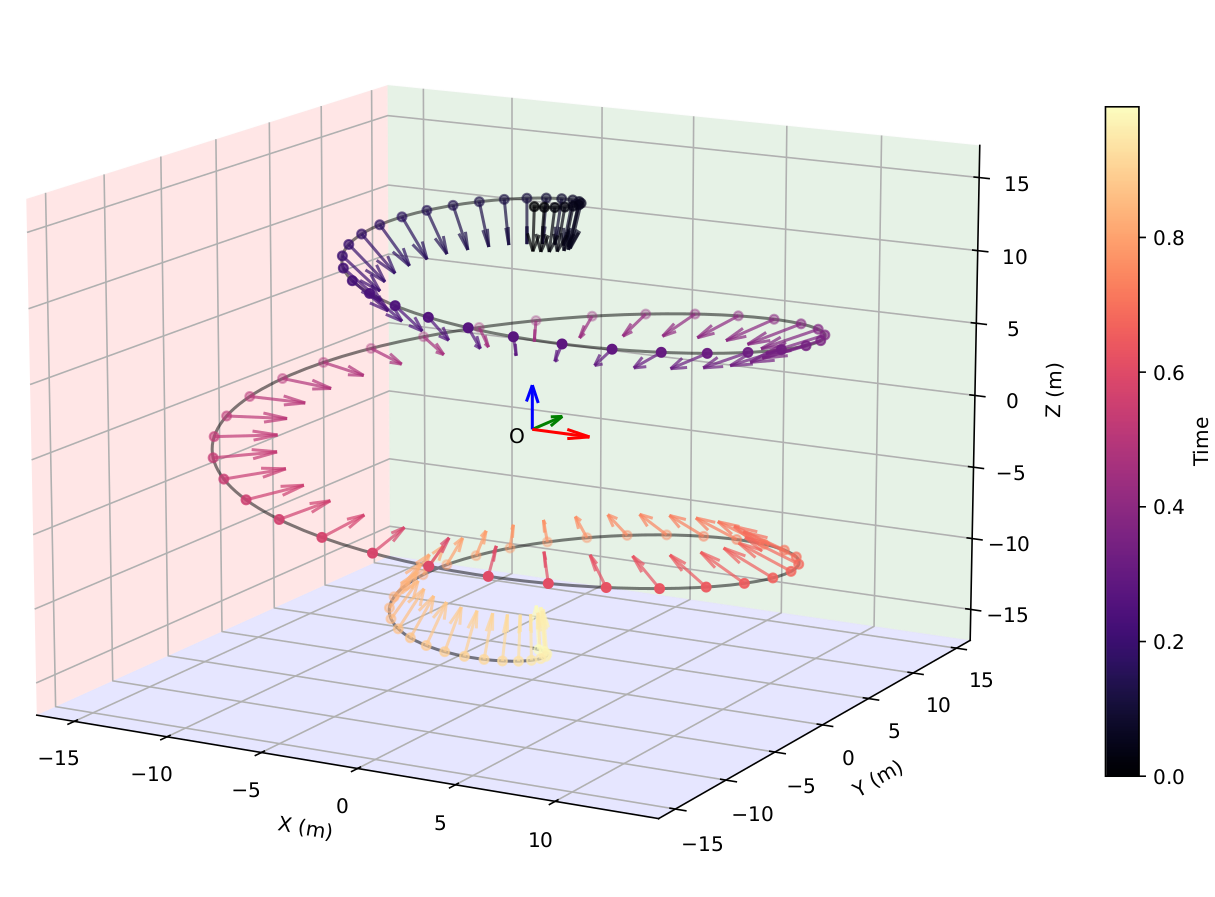}
         \caption{}
         \label{fig:trajectory}
     \end{subfigure}
        \caption{(a) \textbf{Dataset objects}: ESA science fleet spacecraft models used to generate the dataset. (b) \textbf{Test Trajectory}: Visualization of the camera trajectory relative to the target spacecraft, whose body frame is denoted by the axes at $O$. The arrows show the boresight direction at each point.}
        \label{fig:dataset_vis}
        \vspace{-1.0\baselineskip}
\end{figure*}
\subsection{Data Generation}

We test the performance of our method on a custom dataset of image sequences of 10 space objects from ESA's science fleet\footnote{\url{https://scifleet.esa.int/}} as shown in Fig.~\ref{fig:dataset_vis}. For consistency, we simulate the same relative trajectory for all the spacecraft models and the perspective camera model. The simulated relative motion is a spherical spiral as shown in Fig.~\ref{fig:dataset_vis} around a sphere of radius 16m. The trajectory allows for spacecraft to incrementally reveal new views and to have varying photometric shifts between time steps along the trajectory. We note that there are no limitations on the nature of this trajectory, as long as the spacecraft is sufficiently resolved in the image and the photometric shifts between two subsequent frames are not large and abrupt. We simplify the data generation process by scaling down the spacecraft models appropriately to fit the FOV, so we can test in under the same trajectory. Further, we simplify the spacecraft's appearance by reducing the material specularity. Since we ignore the view-based appearance model in~\acrshort{3dgs}, our method in its current form cannot deal with flares and sharp specular reflections. We use Nvidia Omniverse Isaac Sim with the ray-tracing render engine to generate a sequence of 1000 synthetic images along the trajectory for each object. Through the replicator module, we generate an RGB image, an aligned depth image, a segmentation image, and the relative ground-truth pose for each time step of the trajectory. During data-loading, we add random noise in the color and depth images with a standard deviation of 0.2 (intensity) and 0.025m respectively. Furthermore, we add edge bleeding artefacts in the depth image to emulate a stereo depth output.


\begin{figure*}
    \centering
    \includegraphics[width=\linewidth]{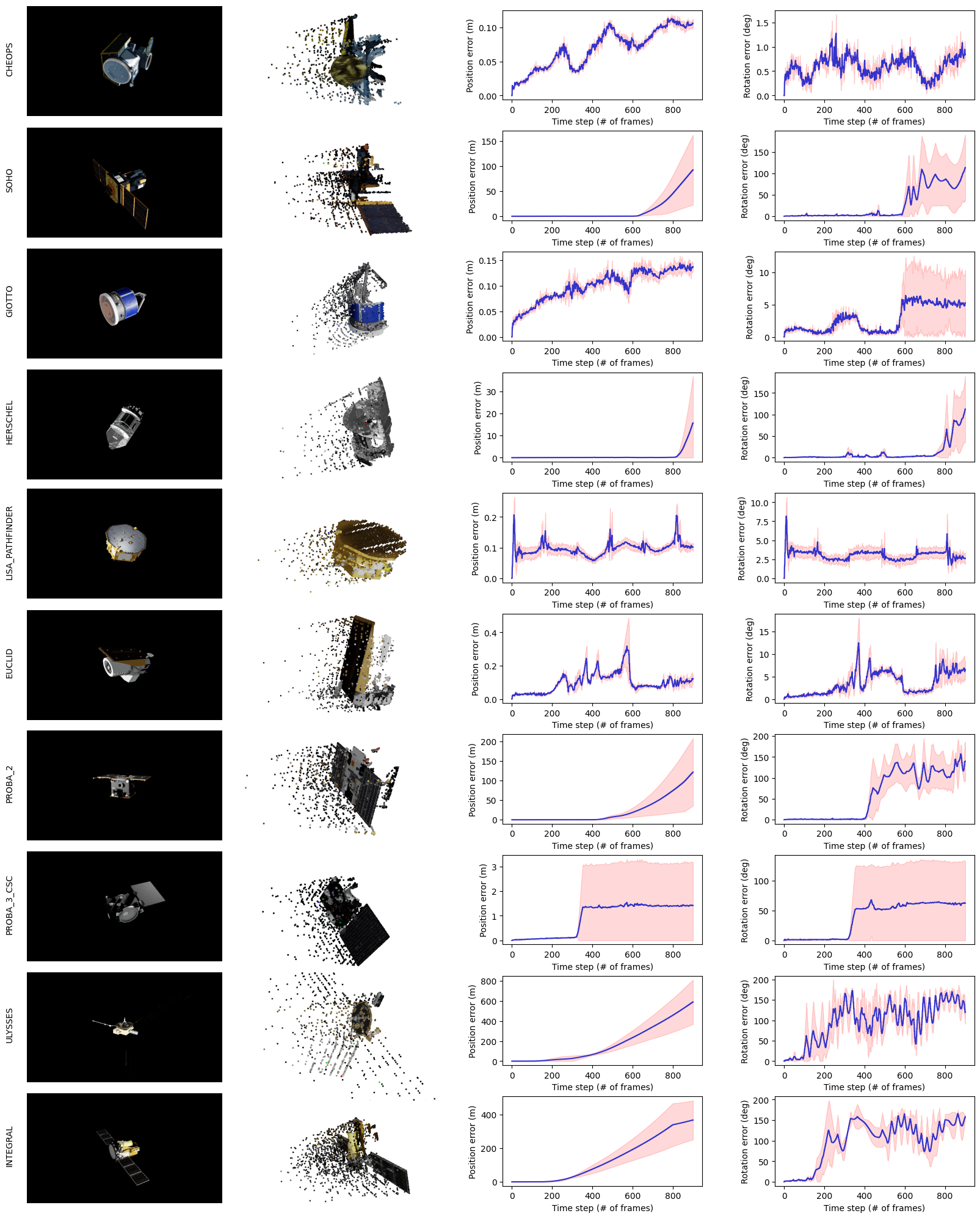}
    \caption{\textbf{Results}: Pose tracking errors over 1000 frames for 10 synthetic spacecraft models of diverse geometry and textures. A sample input image and the point cloud of the first RGB-D frame for each spacecraft are shown on the left.}
    \label{fig:results_tracking}
    \vspace{-1.0\baselineskip}
\end{figure*}

\subsection{Evaluation}
We evaluate the performance of our approach by running optimization for reconstruction and tracking in alternate frames. The reconstruction performance is measured by the bi-directional chamfer distance between the ground-truth surface point clouds and the point cloud obtained from the online reconstruction. The chamfer distance metric quantifies the similarity between two point clouds by averaging the pair-wise distances of the closest points. This is given by:
\begin{equation}
    \text{CD}(\mathbf{x}_1, \mathbf{x}_2) = \frac{1}{2n_1} \sum_{i=1}^{n_1} |x^i_1 - x^{i*}_{2}| + \frac{1}{2n_2} \sum_{j=1}^{n_2} |x^j_2 - x^{j*}_{1}|
\end{equation}
where, $x^{i*}$ and $x^{j*}$ are the nearest neighbour points in the other point cloud. We compute chamfer distance over 20000 points by uniformly sampling from each point cloud.

The tracking error is simply computed as the translation error and rotation error to the annotated ground truth pose:
\begin{align}
    t_{err} \, &=\, || \mathbf{t}_{CO} - \mathbf{t}_{CO}^* ||_2 \; \nonumber \\ 
    \theta_{err} &= 2 \, \arccos (\mathbf{q_{CO}} \otimes \mathbf{\Bar{q}_{CO}^*})_w
\end{align}
where, $\mathbf{t}_{CO}^*$ and $\mathbf{t}_{CO}^*$ are ground truth position and orientation.

\begin{table}[]
    \centering
    \caption{Chamfer distance evaluation}
    \begin{tabular}{p{10mm} p{13mm} p{13mm} p{13mm} p{13mm}}
    Spacecraft & CD@0 & CD@250 & CD@500 & CD@1000 \\ \hline\hline \\
       CHEOPS  &1.18$\pm$0.02 & 0.61$\pm$0.02 & 0.54$\pm$0.02 & \textbf{0.54$\pm$0.02} \\ \hline
       SOHO  &1.96$\pm$0.00 & 1.53$\pm$0.03 & \textbf{1.04$\pm$0.14} & 1.27$\pm$0.16 \\ \hline
       Giotto   &0.62$\pm$0.01 & 0.05$\pm$0.01 & 0.02$\pm$0.00 & \textbf{0.01$\pm$0.00}\\ \hline
       Herschel & 0.99$\pm$0.01 & 0.34$\pm$0.01 & 0.26$\pm$0.01 & \textbf{0.22$\pm$0.00}\\ \hline
       LISA Pathfinder & 1.04$\pm$0.01 &  0.57$\pm$0.01 & 0.50$\pm$0.01 &\textbf{0.49$\pm$0.01}\\ \hline
       Euclid  & 0.61$\pm$0.01 & 0.19$\pm$0.01 & 0.17$\pm$0.00 & \textbf{0.16$\pm$0.01}\\ \hline
       Proba 2  & 0.78$\pm$0.01 &0.45$\pm$0.01 &\textbf{0.38$\pm$0.01} &0.39$\pm$0.01\\ \hline 
       Proba 3 CSC  &1.43$\pm$0.01 & 1.11$\pm$0.02 & 0.94$\pm$0.12 & \textbf{0.83$\pm$0.08} \\ \hline
       Ulysses & \textbf{9.94$\pm$0.01} & 31.97$\pm$1.82 & 34.60$\pm$1.31 & 33.80$\pm$1.57 \\\hline
       Integral & 2.35$\pm$0.01 & 1.96$\pm$0.01 & 1.53$\pm$0.16 &  \textbf{1.23$\pm$0.27}\\
    \end{tabular}
    \label{tab:recon_results}
    \vspace{-1.0\baselineskip}
\end{table}

\section{Results and Discussion}
While the reconstruction and tracking performance are coupled in our method, it is helpful to analyze their evolution over the trajectory separately at first. Table~\ref{tab:recon_results} shows the variation of the bi-directional chamfer distance between the reconstructed point cloud and the reference point cloud for each object. As newer views of the target object are revealed, the proposed method generally leads to consistent improvement in the accuracy and completeness of the reconstructed 3D model. The convergence of the reconstruction error is sensitive to the pose tracking error. When new 3D Gaussians are added, the last estimated relative pose is used to project them from the camera frame to the object frame. Consequently, larger pose tracking errors can lead to diverging reconstruction errors. This can be seen in the case of SOHO, Proba 2, and Ulysses models where the chamfer distance at the last step is not the lowest. In the case of Ulysses, the reconstruction errors present a clear sign of early tracking drift leading to sharp divergence. Another notable observation is the much lower improvement in error between the 500$^{th}$ and the 1000$^{th}$ step. Depending on the spacecraft geometry, most regions of the spacecraft can be revealed by step 500. This effect is dominated by stray points in the reconstruction far away from the local surfaces of the object, resulting from the noise in the depth images. This effect can be observed in the initial point cloud shown in Fig.~\ref{fig:results_tracking} as well as the final reconstructions shown in Fig. 1. 

The tracking results for each of the spacecraft in terms of translation and rotation error are shown in Fig.~\ref{fig:results_tracking}. For clarity, we separate the tracking error for the first 500 steps from that of the entire 1000-step sequence. We outline three direct observations. First, our method allows accurate tracking of diverse target objects over short durations. Second, over 500 steps, the tracking error for 6/10 spacecraft models is maintained under the tight bounds of 0.5 m in translation and 10 degrees in rotation. Finally, over the entire 1000 steps, the tracking error for 6/10 objects diverges uncontrollably, while that of 4/10 objects remains within a reasonable error bound. Together with the reconstruction results, these observations provide the preliminary validation of our incremental approach using 3D Gaussian representations to track unknown objects. On a system with 12th Gen Intel Core i7-12800H CPU and Nvidia RTX 3080 Ti Laptop GPU, it takes approximately 1.4s and 1.1s to optimize the reconstruction (120 steps) and pose (80 steps) respectively.    

The experiments demonstrate that the proposed approach and its application to unknown spacecraft tracking are effective. However, the current implementation is limited in robustness and practicality by several factors that can be addressed in future works. The foremost limitation of the approach is the tracking drift over a longer duration. The key factor behind this is the lack of a global pose optimization scheme. Since the pose optimization only considers a single frame, any tracking error cascades negatively. Remember that we use the latest pose available in the previous iteration for optimizing the object parameters and adding newer points. While the repeated reconstruction compensates the tracking error to a certain extent, it diverges beyond a certain margin of error. We observe two main sources that cause this error: (1) Visual ambiguity in regions with low-intensity values and (2) view-dependent changes in appearance. The first source is evident from the tracking results of Integral, Proba-2, and Proba-3-CSC spacecraft which have large regions of surfaces that are dark and do not provide enough contribution to the loss. The tracking drift starts rising noticeably when large parts of the spacecraft observable in the image have close to zero intensity. The second factor is observed in the case of Ulysses where the simulated images contain sharp view-dependent artefacts on the long booms of the spacecraft from data generation.

The improvement of the loss function to deal with low-intensity regions can alleviate abrupt tracking errors seen in the experiments. On the other hand, better keyframe management and pose graph optimization can solve the long-term tracking drift resulting from accumulating errors. Further, naive extrapolation of the relative pose used directly to reconstruct the object and add more Gaussians limits the magnitude of pose shift that can be handled between two consecutive images. From a design perspective, the requirement of depth images can be relaxed by initializing points with a high variance around the existing points based on the color image only. The requirement of a segmentation mask at each frame may also be relaxed by only taking the segmentation mask in the first image and then using the rendered image to mask the future frames. An important factor to include in future work is the efficient representation of specularity, which may be done using spherical harmonics coefficients for the Gaussians. The computation time can be improved by using a second-order optimizer. Finally, the analysis can be extended to a comprehensive range of relative trajectories including active maneuvers to assess the generalizability of our approach.














\section{Conclusions}
This work proposes a novel approach for incremental reconstruction and pose tracking of unknown dynamic objects using 3D Gaussian representation. It assumes no prior knowledge of the object or its motion and requires no pre-training or neural elements. Relying on a differentiable rendering framework that is fast, it is more suitable for online applications than other contemporary methods. The effectiveness of the method is validated on the task of tracking an unknown dynamic spacecraft. On a custom dataset of ten spacecraft with diverse geometry and appearance, the tracking accuracy over shorter duration is less than 0.5m in translation and 10 deg in rotation. The aspects of longer duration tracking drift are discussed along with recommendations for improvements, paving the way forward for generalizable object-centric perception in space robotics.
\section*{Acknowledgments}

This work is supported by the Fonds National de la Recherche (FNR) Industrial Fellowship grant (15799985) and Redwire Space Europe.

\small
\bibliographystyle{IEEEtran.bst}
\bibliography{ref}
\end{document}